\documentclass[runningheads]{llncs}
\usepackage[T1]{fontenc}
\usepackage{graphicx} % For including figures
\usepackage{amsmath}  % For math formatting
\usepackage{amsfonts} % For math fonts
\usepackage{amssymb}  % For additional math symbols
\usepackage{enumitem}
\usepackage{pdflscape}
\usepackage{amsmath}
\usepackage{hyperref}
\usepackage{graphicx}
\begin{document}

\title{Feature Fusion for Improved Classification: Combining Dempster-Shafer Theory and Multiple CNN Architectures}
%
%\titlerunning{Abbreviated paper title}
% If the paper title is too long for the running head, you can set
% an abbreviated paper title here
%
\author{Ayyub Alzahem\inst{1}\orcidID{0000-0003-0274-8808} \and
Wadii Boulila\inst{1}\orcidID{0000-0003-2133-0757} \and
Maha Driss\inst{1}\orcidID{0000-0001-8236-8746} \and
Anis Koubaa\inst{1}\orcidID{0000-0003-3787-7423} }

\institute{Robotics and Internet-of-Things Laboratory, Prince Sultan University, Riyadh 12435, Saudi Arabia}
\authorrunning{A. Alzahem et al.}
% First names are abbreviated in the running head.
% If there are more than two authors, 'et al.' is used.
%
% \institute{Princeton University, Princeton NJ 08544, USA \and
% Springer Heidelberg, Tiergartenstr. 17, 69121 Heidelberg, Germany
% \email{lncs@springer.com}\\
% \url{http://www.springer.com/gp/computer-science/lncs} \and
% ABC Institute, Rupert-Karls-University Heidelberg, Heidelberg, Germany\\
% \email{\{abc,lncs\}@uni-heidelberg.de}}
%
\maketitle              % typeset the header of the contribution
\begin{abstract}
Addressing uncertainty in Deep Learning (DL) is essential, as it enables the development of models that can make reliable predictions and informed decisions in complex, real-world environments where data may be incomplete or ambiguous. This paper introduces a novel algorithm leveraging Dempster-Shafer Theory (DST) to integrate multiple pre-trained models to form an ensemble capable of providing more reliable and enhanced classifications. The main steps of the proposed method include feature extraction, mass function calculation, fusion, and expected utility calculation. Several experiments have been conducted on CIFAR-10 and CIFAR-100 datasets, demonstrating superior classification accuracy of the proposed DST-based method, achieving improvements of 5.4\% and 8.4\%, respectively, compared to the best individual pre-trained models. Results highlight the potential of DST as a robust framework for managing uncertainties related to data when applying DL in real-world scenarios.

\end{abstract}
\section{Introduction}
\label{sec:introduction}

The exponential increase in data size over recent years has become a rich foundation for innovation, particularly in the field of Deep Learning (DL), which thrives on large datasets. As data volumes continue to expand, DL techniques are yielding increasingly impressive results, underscoring their effectiveness in harnessing the power of massive datasets \cite{griffiths2023bayes,li2023gaussian}. Currently, pre-trained DL models are required due to the substantial computational and temporal resources needed for training DL models from scratch. Pre-trained models, developed on comprehensive and heterogeneous datasets, provide a foundational base for fine-tuning specific tasks, thereby considerably economizing on training duration and resources. However, the data are not always accurate and are often marred by ambiguities, uncertainties, and incomplete knowledge \cite{boulila2017sensitivity,ferchichi2017propagating,ferchichi2018reducing}. Numerous factors, including noise in the data, missing or incomplete values, and the intrinsic randomness of the underlying processes, contribute to the uncertainty that arises. Moreover, given that pre-trained DL models are constructed and trained with varied data, each model offers a unique perspective. Consequently, amalgamating the insights from several pre-trained DL models can address complex problems more effectively than any singular model could.
In literature, many approaches have been developed to deal with uncertainty by combining information from different sources. Among the approaches is data fusion, a process that integrates multiple sources of data to produce more consistent, accurate, and useful information than that provided by any individual data source \cite{varone2024finger}. Data fusion plays a crucial role in various fields, such as remote sensing, surveillance, and medical diagnostics, where making informed decisions based on comprehensive data analysis is essential. Dempster-Shafer Theory (DST) is an important data fusion method that handles ambiguity and partial ignorance \cite{ZHU2021306,TONG2021275}. DST provides a solid framework for handling uncertainty, including the Frame of Discernment, Basic Probability Assignment (BPA), and Dempster's rule of combination.

This paper presents an approach based on the DST method to combine several pre-trained DL model decisions. Using the strong structure of the DST, our algorithm aims to integrate all the viewpoints and information found in each model, paving the way for more reliable and accurate classifications \cite{10388220}. The algorithm is expressed in a sequence of logical steps that include feature extraction, mass function construction, masses fusion, and, at the end, estimating the utility of predictions.

The motivation behind this work is to leverage the strengths of DST in providing a refined representation of uncertainty \cite{ZHU2021306,TONG2021275}. Furthermore, our goal is to harness the collective intelligence embedded within multiple pre-trained models \cite{math10214030}. The synergy of DST and pre-trained models is explored through meticulous experiments conducted on CIFAR-10 and CIFAR-100 datasets. The paper attempts to provide a comprehensive explanation of the suggested methodology and the evaluations carried out to reveal the potential of the DST-based ensemble method in improving the performance of pre-trained models.

The remainder of this document is structured as follows: Section \ref{sec:background} provides a detailed exposition of the DST and its significance in uncertainty quantification. Section \ref{sec:literature} delves into the relevant literature encompassing pre-trained models and ensemble methods. Section \ref{sec:proposed} elucidates the proposed algorithm, delineating each phase of the methodology. Section \ref{sec:results} furnishes a comprehensive account of the evaluations conducted and the results obtained, which are further discussed in Section \ref{sec:discussion}. Finally, Section \ref{sec:conclusion} encapsulates the key takeaways from this work and ponders upon the prospective directions for future research. Following this, Section \ref{sec:acknowledgement} extends gratitude to the individuals and institutions that contributed to this work.

\section{Background}
\label{sec:background}

Dempster-Shafer Theory (DST), also known as the theory of belief functions, is a mathematical framework for modeling uncertainty, evidence, and belief \cite{shafer1992dempster,dempster1968generalization}. DST provides a way to combine evidence from different sources to calculate the probability of an event. Unlike traditional probability theory, which requires precise probabilities, DST allows for the representation of uncertainty through belief functions, offering a more flexible approach to handle incomplete and ambiguous information. It assigns degrees of belief to a set of hypotheses rather than to individual events, enabling the combination of partial knowledge and evidence to support decision-making processes \cite{LUO2024102070,XUE2024121285,CHEN2024114172}. The main concepts of DST theory are the following:

\paragraph{Frame of Discernment}
The Frame of Discernment denoted as \( \Theta \), is pivotal in DST, representing an exhaustive and mutually exclusive collection of all potential outcomes or hypotheses. For instance, in a binary classification scenario, \( \Theta \) would comprise two elements, each symbolizing one class. 

\paragraph{Basic Probability Assignment (BPA)}

Assigned to each subset of \( \Theta \), including the empty set, the BPA, \( m: 2^{\Theta} \rightarrow [0, 1] \), indicates the mass or amount of belief exclusively allotted to that subset.

\begin{equation}
m(\emptyset) = 0 \quad \text{and} \quad \sum_{A \subseteq \Theta} m(A) = 1
\end{equation}

\paragraph{Belief and Plausibility}
Derived from BPA are two pivotal measures:
\begin{itemize}
    \item \textbf{Belief Function (Bel)}: This is a representation of total evidence favoring a subset \( A \) of \( \Theta \).
    
    \begin{equation}
    \text{Bel}(A) = \sum_{B \subseteq A} m(B)
    \end{equation}

    \item \textbf{Plausibility Function (Pl)}: Serving as an upper bound on belief, it gauges the potential evidence favoring subset \( A \) considering all subsets.
    
    \begin{equation}
    \text{Pl}(A) = \sum_{B \cap A \neq \emptyset} m(B)
    \end{equation}
\end{itemize}

\paragraph{Doubt and Commonality}
In tandem with Belief and Plausibility, DST introduces:
\begin{itemize}
    \item \textbf{Doubt}: Representing the complement of Plausibility, it reflects the cumulative belief negating set \( A \).
    
    \begin{equation}
    \text{Doubt}(A) = 1 - \text{Pl}(A)
    \end{equation}

    \item \textbf{Commonality Function (q)}: Quantifying the cumulative belief endorsing set \( A \), either directly or indirectly.
    
    \begin{equation}
    q(A) = \sum_{B \subseteq A} m(B)
    \end{equation}
\end{itemize}

\paragraph{Combination Rule}
Dempster's rule of combination encapsulates the essence of DST's evidence fusion capabilities. For two BPAs, \( m_1 \) and \( m_2 \), the resultant combined mass function, \( m_{12} \), is calculated as:

\begin{equation}
m_{12}(A) = \frac{1}{1 - K} \sum_{B \cap C = A} m_1(B) \times m_2(C)
\end{equation}

Here, \( K \) quantifies the conflict between \( m_1 \) and \( m_2 \):

\begin{equation}
K = \sum_{B \cap C = \emptyset} m_1(B) \times m_2(C)
\end{equation}

When \( K \) is significant, it implies substantial conflict, necessitating caution, whereas a smaller \( K \) denotes lesser conflict and more reliability in combined evidence.

The DST's advanced handling of uncertainty makes it a solid foundation for many real-world applications. This is especially true in situations where there is conflicting or incomplete evidence.

\section{Literature Review}
\label{sec:literature}
This section explores different cutting-edge frameworks and approaches that represent the convergence of the DST and DL paradigms.

Yaghoubi et al. \cite{yaghoubi2022cnn} delineated a unique ensemble DL framework named CNN-DST, meticulously crafted to scrutinize structural faults leveraging vibration data. This framework represents a fusion of Convolutional Neural Networks (CNN) and DST, showcasing a synergy where multiple CNNs undergo rigorous training, and their outputs are seamlessly integrated through an enhanced DST method. The efficacy of the CNN-DST framework was rigorously evaluated in a detailed assessment involving a cohort of turbine blades, where it achieved a 97.19\% prediction accuracy. This performance not only surpassed that of its counterparts but also established a high benchmark in the realm of structural fault detection.

Lateef et al. \cite{qiu2022hybrid} proposed a novel intrusion detection paradigm that utilizes traffic flows and their initial packets as the basis for analysis. By applying DST to combine flow-based and packet-based IDS predictions, this hybrid system is designed to enhance both the accuracy and efficiency of intrusion detection. The fusion mechanism adeptly navigates through the uncertainty inherent in classification tasks and establishes a robust defense against a wide range of cyber-attacks, thereby setting a high benchmark in the domain of cybersecurity.

Tong et al. \cite{tong2021evidential} introduced a classifier that combines DST with a CNN, pioneering the domain of set-valued classification. This evidential DL classifier extracts high-dimensional features. These features are then transformed into mass functions. The novel end-to-end learning strategy highlights its potential in image recognition, signal processing, and semantic-relationship classification tasks.

Peñafiel et al. \cite{penafiel2020applying} introduced a new methodology for developing AI systems designed to enhance decision-making processes. Their method, deeply rooted in Dempster-Shafer's plausibility theory, crafts mass assignment functions (MAF) that encode expert knowledge. The validation of this approach shows its potential to generate accurate classifications and insightful explanations.

Tian et al. \cite{tian2020deep} explored a hybrid intrusion detection system built on the foundation of DST and DL. By utilizing DST to integrate predictions from various IDS, the system demonstrates a promising approach to improving the accuracy and efficiency of intrusion detection, thereby strengthening defenses against a wide array of cyber threats.

Advancing the frontier of feature selection, Zheng et al. \cite{zheng2019feature} proposed a novel ensemble learning algorithm named TSDS, which is an offspring of Tsallis entropy and DST. This algorithm creates a robust mechanism for selecting relevant features and discarding redundant ones through an enhanced correlation criterion and a forward sequential approximate Markov blanket. TSDS was subjected to several tests across various datasets from diverse domains. The results highlight its effectiveness in significantly enhancing classification performance.

Denoeux \cite{denoeux2019logistic} presented a detailed exploration of logistic regression and its nonlinear counterparts, revealing how these classifiers can transform input or higher-level features into Dempster-Shafer mass functions. This investigation highlights the capability of mass functions to provide a detailed understanding of evidence, differentiating between the lack of evidence and conflicting evidence. As a result, it offers a new perspective on the importance of input features in logistic regression and on interpreting the outputs of hidden units in multilayer neural networks.

\begin{table}[!ht]
\centering
\scriptsize
\caption{Review of works discussing DST and DL for fault detection and classification.}
\label{tab:relates_works}
\begin{tabular}{p{2cm}cp{3cm}p{3cm}p{3cm}} 
 Ref. & Year & Methodology & Findings & Limitations \\ \hline
Yaghoubi et al. \cite{yaghoubi2022cnn} & 2022 & Ensemble DL (CNN-DST) & 97.19\% prediction accuracy & Structure design and hyperparameter tuning challenges \\ \hline

Tong et al. \cite{tong2021evidential} & 2021 & DS theory and CNN for set-valued classification & Improved classification accuracy & Overfitting or underfitting, varying performance based on training data quality \\ \hline

Zheng et al. \cite{zheng2019feature} & 2019 & Tsallis entropy and Dempster–Shafer evidence theory (TSDS) & Effective feature selection & - \\ \hline

Denoeux \cite{denoeux2019logistic} & 2019 & Logistic Regression, Neural Networks and DST & Insight into role of input features & - \\ \hline

Lateef et al. \cite{qiu2022hybrid} & 2022 & Hybrid intrusion detection system (DST) & Enhanced accuracy and efficiency in intrusion detection & Dependent on quality and quantity of training data, model complexity \\ \hline

Peñafiel et al. \cite{penafiel2020applying} & 2020 & Dempster-Shafer’s Plausibility Theory based classification & Accurate classifications and explanations & - \\ \hline

Tian et al. \cite{tian2020deep} & 2020 & Hybrid intrusion detection (DST and DL) & Improved accuracy and efficiency in intrusion detection & - \\ \hline

\end{tabular}
\end{table}

\section{Proposed Algorithm}
\label{sec:proposed}

\begin{figure*}[h!]
\centering
\includegraphics[width=\textwidth]{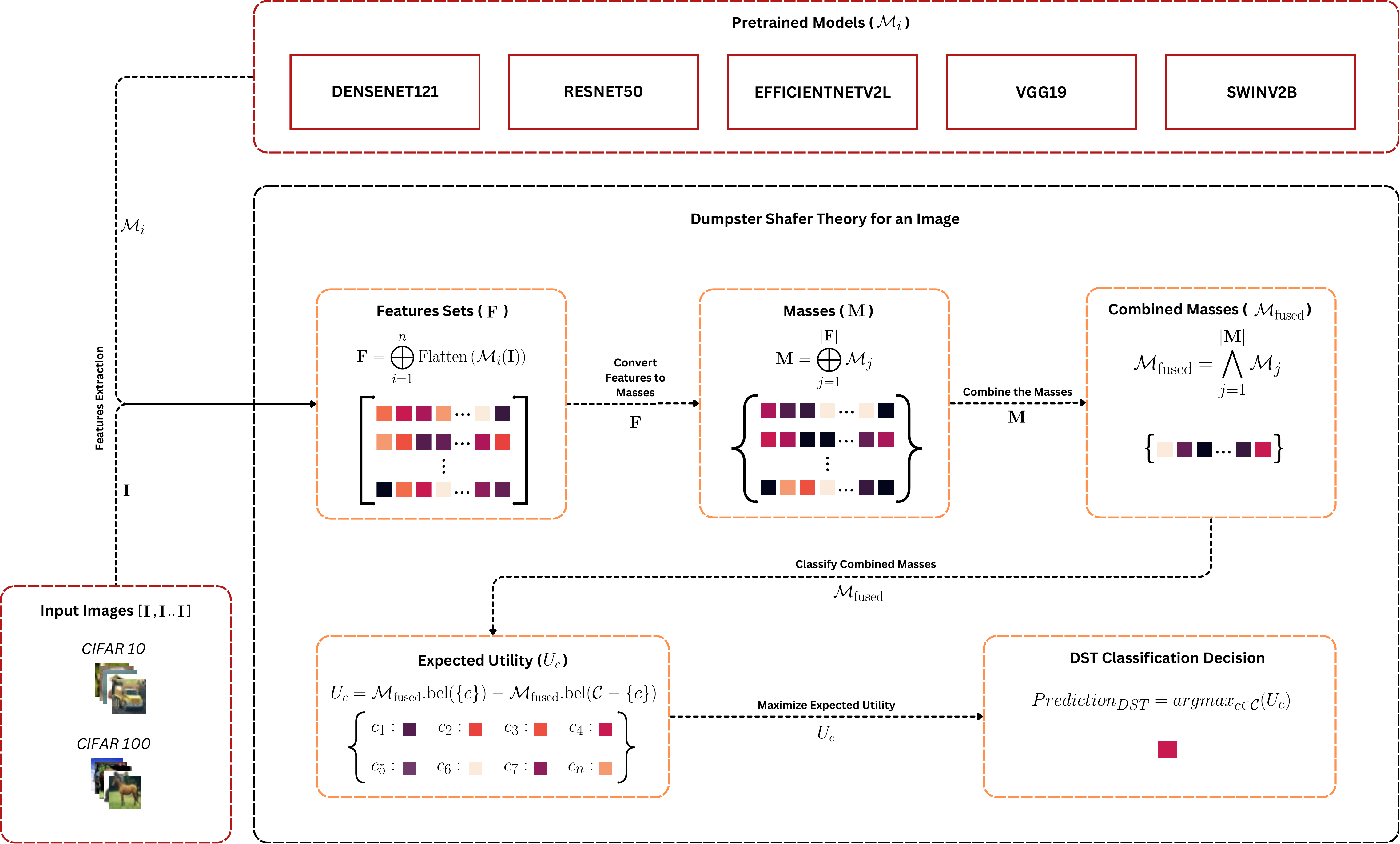}
\caption{Illustration of the Proposed Approach.}
\label{fig:proposed_algorithm}
\end{figure*}

In several recent works in ML, there has been a significant rise in the use of pre-trained models. Although these models offer distinct benefits, they also come with their own set of limitations. This work introduces a novel algorithm based on the DST that seamlessly integrates the features of multiple pre-trained models. Through DST's robust framework, the algorithm seeks to harness and amalgamate the distinct knowledge embedded within each model, aiming for enhanced classifications. The proposed methodology is articulated in four distinct phases, elaborated below and illustrated in Figure \ref{fig:proposed_algorithm}.

\subsection{Feature Extraction}

The initial phase of the proposed approach involves the extraction of feature vectors. The input image, denoted as $\mathbf{I}$, is processed by each pre-trained model, $\mathcal{M}_{i}$. This processing translates into a set of outputs or features that offer a particular perspective on the image's attributes. These outputs are subsequently reshaped into a flattened 1D tensor to simplify the ensuing calculations. The aggregation of these vectors from diverse models constitutes the comprehensive feature list, $\mathbf{F}$, as articulated in Equation \ref{eq:feature_extraction}.

\begin{equation}
\mathbf{F} = \bigoplus_{i=1}^{n} \text{Flatten}\left(\mathcal{M}_{i}(\mathbf{I})\right)
\label{eq:feature_extraction}
\end{equation}

\subsection{Mass Function Construction}

Once the features are extracted, the ensuing challenge entails encoding the uncertainty inherent in the prediction of each class. At this step, the DST provides an invaluable advantage, offering a more refined representation of uncertainty compared to traditional probability models. Each feature's prediction for a class is subsequently converted into a mass value, as detailed in Equation \ref{eq:mass_definition}.

\begin{equation}
m_c =
\begin{cases}
f_{j,c} & \text{if } f_{j,c} \geq 0.5 \sum_{k=1}^{n} \left| f_{j,k} \right| \\
0 & \text{otherwise}
\end{cases}
\label{eq:mass_definition}
\end{equation}\textbf{}

This sequence of mass values across all classes defines the mass function for each feature, symbolized by $\mathcal{M}_j$. The collation of these values is presented in Equation \ref{eq:mass_aggregation}.

\begin{equation}
\mathcal{M}_j = \bigcup{c=1}^{n} m_c
\label{eq:mass_aggregation}
\end{equation}

Given the pivotal role of mass functions in DST, it is indispensable to validate their structure. A crucial validation step is normalization, ensuring that the cumulative values of each mass function equate to one. The normalization process is depicted in Equation \ref{eq:mass_normalization}.

\begin{equation}
\mathcal{M}_j = \text{normalize}(\mathcal{M}_j)
\label{eq:mass_normalization}
\end{equation}

Following normalization, a consolidated ensemble of all mass functions is constructed, represented as $\mathbf{M}$ in Equation \ref{eq:mass_construction}.

\begin{equation}
\mathbf{M} = \bigoplus_{j=1}^{\left| \mathbf{F} \right|} \mathcal{M}_j
\label{eq:mass_construction}
\end{equation}

\subsection{Mass Function Fusion}

The fusion of individual mass functions aims to create a collective, enriched perspective, encapsulated as $\mathcal{M}_{\text{fused}}$, in line with Equation \ref{eq:mass_fusion}.

\begin{equation}
\mathcal{M}_{\text{fused}} = \bigwedge{j=1}^{\left| \mathbf{M} \right|} \mathcal{M}_j
\label{eq:mass_fusion}
\end{equation}

\subsection{Expected Utility Calculation and Prediction}

The final phase is centered on deriving predictions. Utilizing the fused mass function, the algorithm calculates the expected utility for each class. This utility captures the belief associated with each class based on DST principles. This process is presented in Equation \ref{eq:expected_utility}.

\begin{equation}
U_c = \mathcal{M}_{\text{fused}}.\text{bel}({c}) - \mathcal{M}_{\text{fused}}.\text{bel}(\mathcal{C} - {c})
\label{eq:expected_utility}
\end{equation}

The algorithm finishes by choosing the class that has the highest expected utility, as detailed in the Equation \ref{eq:final_prediction}.

\begin{equation}
\text{Prediction}{\text{DST}} = \text{argmax}_{c \in \mathcal{C}}(U_c)
\label{eq:final_prediction}
\end{equation}

By merging the strengths of different models using this algorithm, we aim to create predictions that are not just strong and dependable but also detailed. This method has the potential for wide-ranging applications.

\section{Evaluation and Results}
\label{sec:results}

The experiments conducted aimed at evaluating the effectiveness of the proposed DST-based ensemble method in comparison with individual pre-trained models on CIFAR-10 and CIFAR-100 datasets. Each model was trained independently on these datasets, and their performance was gauged based on the classification accuracy, which was computed as the ratio of the number of correct predictions to the total number of items in the testing set. The details of these evaluations are illustrated in Table \ref{tab:results}.

\begin{table}[!ht]
\centering
\caption{Classification accuracy of the pre-trained models on CIFAR-10 and CIFAR-100 datasets.}
\label{tab:results}
\begin{tabular}{p{4cm}p{2cm}p{2cm}}
\hline
Model & \hfil CIFAR-10 & \hfil CIFAR-100 \\
\hline
ResNet50 & \hfil 0.842 & \hfil 0.652 \\
VGG19 & \hfil 0.869 & \hfil 0.655 \\
DenseNet121 & \hfil 0.832 & \hfil 0.621 \\
EfficientNetV2Large & \hfil 0.816 & \hfil 0.628 \\
SwinTransformerV2Big & \hfil 0.879 & \hfil 0.706 \\ \hline
\textbf{DST Ensemble} & \hfil \textbf{0.896} & \hfil \textbf{0.736} \\
\hline
\end{tabular}
\end{table}

\begin{figure*}[!ht]
\centering
\includegraphics[width=3in]{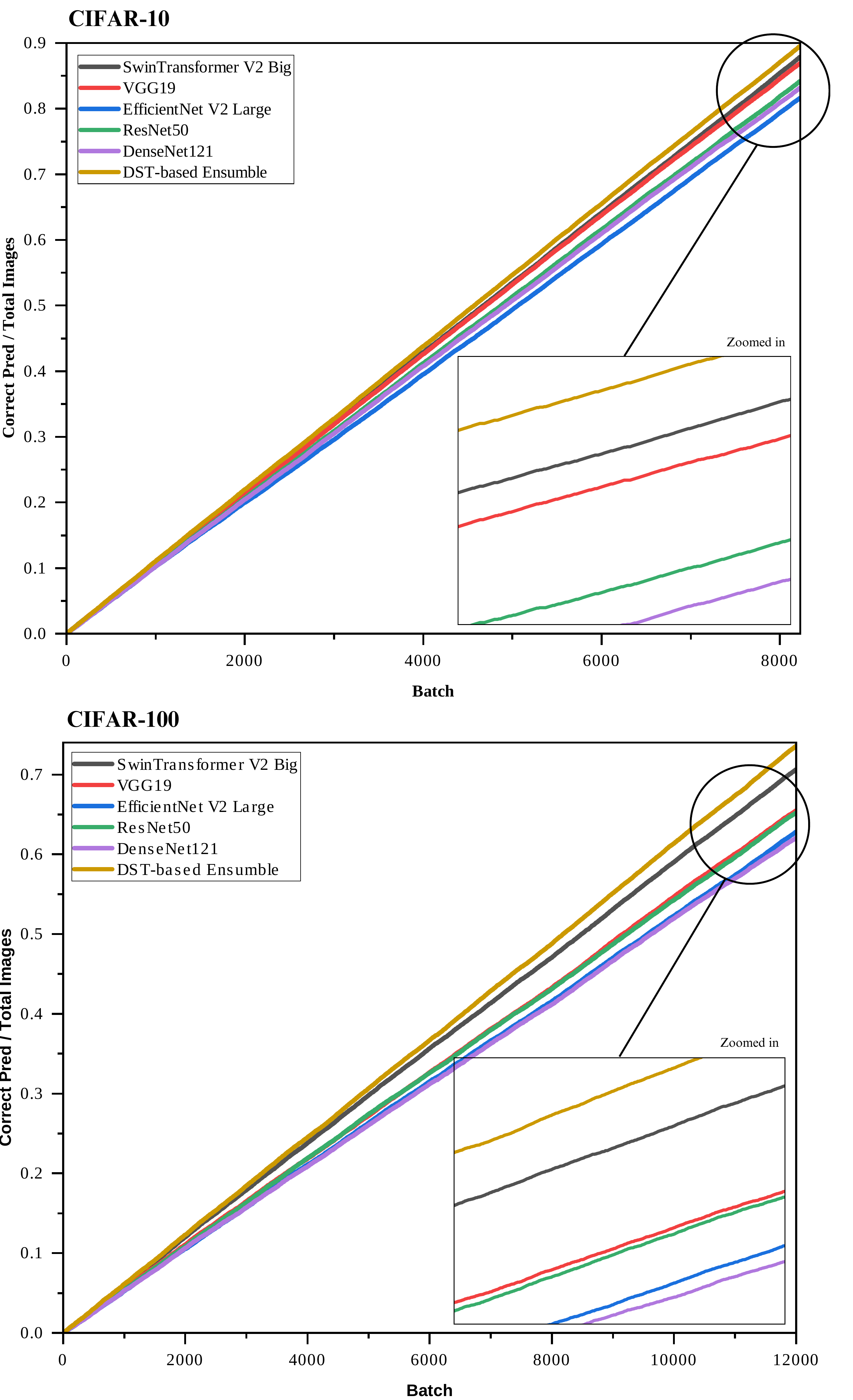}
\caption{Prediction accuracy of pre-trained models and DST-based ensemble on CIFAR-10 and CIFAR-100. A zoomed-in section at the end of the distribution lines reveals the superiority of the DST-based ensemble in achieving higher accuracy.}
\label{fig:accuracy}
\end{figure*}

Figure \ref{fig:accuracy} depicts the prediction accuracy of the individual models and the DST-based ensemble on both CIFAR-10 and CIFAR-100 datasets. A clear advantage of the DST-based ensemble over the individual models is illustrated, emphasizing the ensemble's capability to amalgamate the strengths of diverse architectures to achieve superior performance.

\begin{figure*}[h!]
\centering
\includegraphics[width=3in]{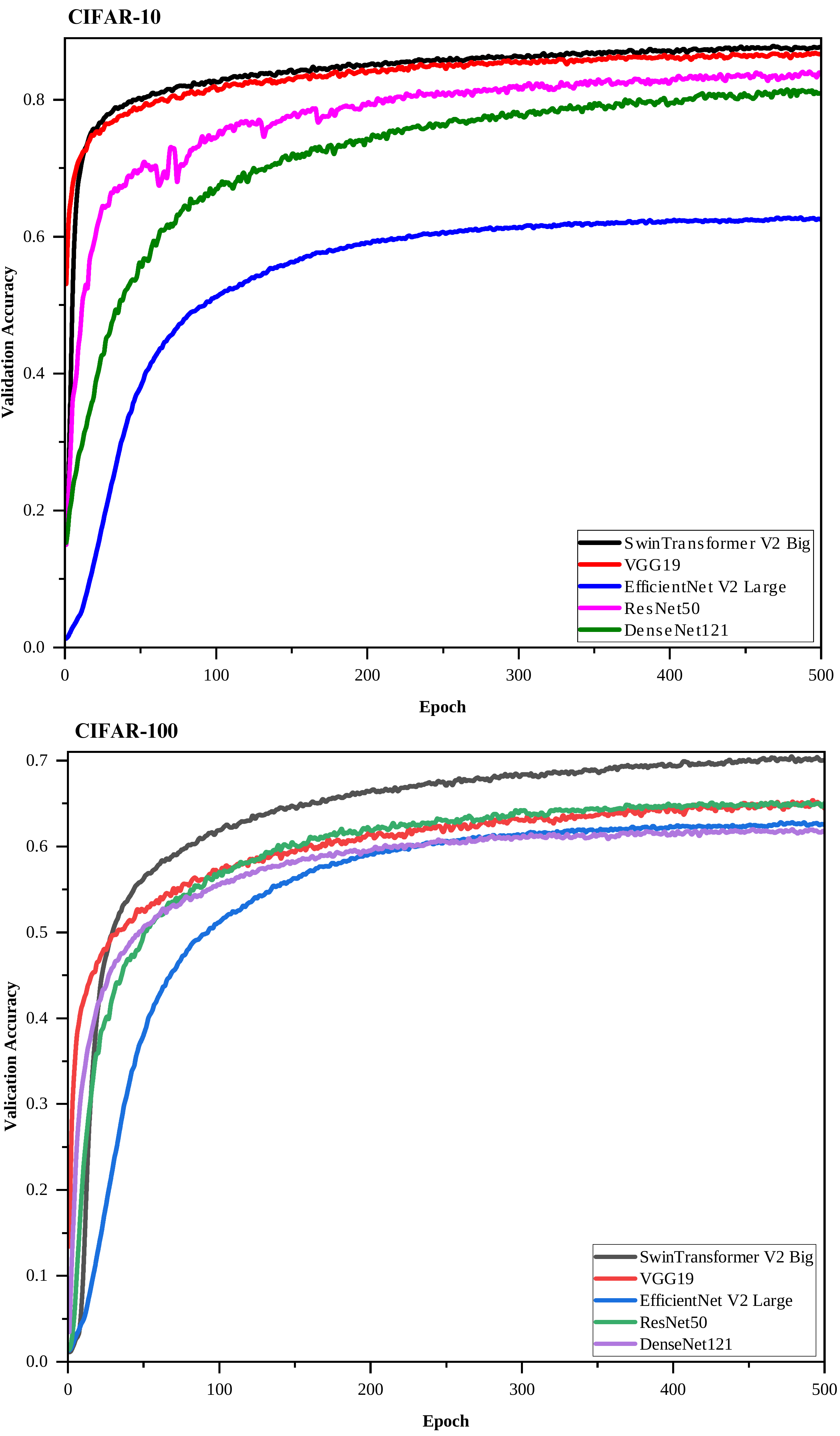}
\caption{Training progress of pre-trained models on CIFAR-10 and CIFAR-100. The subfigures indicate the validation accuracy over epochs, showcasing the comparative performance dynamics of the models.}
\label{fig:progress}
\end{figure*}

The training progress of the pre-trained models on both datasets is elucidated in Figure \ref{fig:progress}. A notable distinction in the validation accuracy progression is observed between SwinTransformerV2Big and the other models, especially on the CIFAR-100 dataset, underpinning the superior generalization performance of SwinTransformerV2Big.

Tables \ref{tab:validation_metrics_cifar10} and \ref{tab:validation_metrics_cifar100} encapsulate a range of validation performance metrics during training on CIFAR-10 and CIFAR-100 datasets, respectively. These metrics, including accuracy, precision, recall, F1 score, and loss, furnish a holistic understanding of the models' performance, reinforcing the observations made from the accuracy evaluations.

\begin{table}[h!]
\centering
\caption{Validation performance metrics during training on CIFAR-10 dataset.}
\label{tab:validation_metrics_cifar10}
\begin{tabular}{p{4cm}p{1cm}p{1cm}p{1cm}p{1cm}p{1cm}}
\hline
Model & \hfil Acc & \hfil P & \hfil R & \hfil F1 & \hfil Loss \\
\hline
DenseNet121 & \hfil 0.832 & \hfil 0.830 & \hfil 0.832 & \hfil 0.831 & \hfil 0.005 \\
EfficientNetV2Large & \hfil 0.816 & \hfil 0.816 & \hfil 0.816 & \hfil 0.815 & \hfil 0.004 \\
ResNet50 & \hfil 0.842 & \hfil 0.846 & \hfil 0.842 & \hfil 0.841 & \hfil 0.004 \\
SwinTransformerV2Big & \hfil 0.879 & \hfil 0.878 & \hfil 0.879 & \hfil 0.878 & \hfil 0.003 \\
VGG19 & \hfil 0.869 & \hfil 0.869 & \hfil 0.869 & \hfil 0.869 & \hfil 0.003 \\
\hline
\end{tabular}
\end{table}

\begin{table}[h!]
\centering
\caption{Validation performance metrics during training on CIFAR-100 dataset.}
\label{tab:validation_metrics_cifar100}
\begin{tabular}{p{4cm}p{1cm}p{1cm}p{1cm}p{1cm}p{1cm}}
\hline
Model & \hfil Acc & \hfil P & \hfil R & \hfil F1 & \hfil Loss \\
\hline
DenseNet121 & \hfil 0.621 & \hfil 0.622 & \hfil 0.621 & \hfil 0.618 & \hfil 0.012 \\
EfficientNetV2Large & \hfil 0.628 & \hfil 0.628 & \hfil 0.628 & \hfil 0.627 & \hfil 0.014 \\
ResNet50 & \hfil 0.652 & \hfil 0.654 & \hfil 0.652 & \hfil 0.649 & \hfil 0.010 \\
SwinTransformerV2Big & \hfil 0.706 & \hfil 0.708 & \hfil 0.706 & \hfil 0.705 & \hfil 0.008 \\
VGG19 & \hfil 0.655 & \hfil 0.657 & \hfil 0.655 & \hfil 0.652 & \hfil 0.011 \\
\hline
\end{tabular}
\end{table}

The computational environment for these experiments comprised a system with a 12th Gen Intel(R) Core(TM) i7-12700K processor clocked at 3.60 GHz, 32.0 GB RAM, and an Nvidia 3080 Ti 12 GB GPU. The training was executed with a batch size of 128, over 500 epochs, employing a learning rate of 0.0001. The Stochastic Gradient Descent (SGD) optimizer and Cross-Entropy loss function were utilized to ensure convergence and optimal performance.

\subsection{Analysis of Other Combinations}
Various combinations of pre-trained models were also experimented to evaluate the versatility and robustness of the proposed DST-based ensemble method. The combinations ranged from pairing a few models together to integrating almost all the discussed models in a single ensemble.

It was observed that the combination of SwinTransformerV2Big, VGG19, EfficientNetV2Large, ResNet50, and DenseNet121 achieved the highest accuracy of approximately 0.896 on CIFAR-10 and 0.736 on CIFAR-100 among all other combinations. Other noteworthy combinations include the ensemble of SwinTransformerV2Big, ResNet152, EfficientNetV2Large, and DenseNet121 which yielded an accuracy of around 0.883 on CIFAR-10 and 0.729 on CIFAR-100, and the ensemble of SwinTransformerV2Tiny, ResNet50, EfficientNetB7, and DenseNet201 with an accuracy of 0.879 on CIFAR-10 and 0.701 on CIFAR-100.

The combination of SwinTransformerV2Tiny, SwinTransformerV2Big, ResNet50, ResNet152, VGG16, VGG19, EfficientNetB7, EfficientNetV2Large, DenseNet121, and DenseNet201 showcased an accuracy of 0.895 on CIFAR-10 and 0.732 on CIFAR-100, indicating that integrating a wide variety of models can potentially lead to improved performance. However, it also indicates a point where the benefits start to decrease, as the improvement in accuracy doesn't match the large increase in the number of models combined.

\section{Discussion}
\label{sec:discussion}

The implementation of the DST-based ensemble method for combining decisions from various pre-trained models has significantly enhanced classification accuracy on the CIFAR-10 and CIFAR-100 datasets. The proposed approach effectively harnesses the collective strengths of individual models, leading to a marked improvement in performance compared to using any single model. The experiments outlined in the previous section confirm the efficacy of the DST ensemble method, showcasing its superiority in terms of classification accuracy.

One of the key insights from this study is the relationship between the size of the ensemble and the achievable accuracy. The optimal combination of models, as demonstrated by the highest accuracy scores, suggests that there is an ideal number of models to be integrated to maximize performance. This observation paves the way for future investigations into finding the most effective model combinations for various tasks, which could further refine and enhance the performance of ensemble learning methods in ML applications.

Moreover, the results underscore the utility of using DST as an ensemble learning framework. By enabling the integration of decisions from multiple DL models, the DST-based approach not only improves accuracy but also introduces a methodological advancement in how ensemble learning can be applied to complex classification tasks. This finding is particularly relevant for the ML community, as it opens new avenues for research and application, particularly in domains where accuracy and reliability are paramount.

\section{Conclusion}
\label{sec:conclusion}
In this study, we proposed a novel method for combining various pre-trained deep learning (DL) models using Dempster-Shafer Theory (DST). The proposed method has significantly improved classification accuracy on the CIFAR-10 and CIFAR-100 datasets, showing accuracy improvements of 5.4\% and 8.4\%, respectively, over the top single models. These results highlight the effectiveness of integrating different models within the DST framework, making it an excellent research topic for future investigations in this area.

\section{Acknowledgment}
\label{sec:acknowledgement}
The authors would like to acknowledge the support of Prince Sultan University
for funding the Article Processing Charges (APC) of this publication.

%
% ---- Bibliography ----
%
% BibTeX users should specify bibliography style 'splncs04'.
% References will then be sorted and formatted in the correct style.
%
\bibliographystyle{splncs04}
\bibliography{mybibliography}
%
% \begin{thebibliography}{8}
% \bibitem{ref_article1}
% Author, F.: Article title. Journal \textbf{2}(5), 99--110 (2016)

% \bibitem{ref_lncs1}
% Author, F., Author, S.: Title of a proceedings paper. In: Editor,
% F., Editor, S. (eds.) CONFERENCE 2016, LNCS, vol. 9999, pp. 1--13.
% Springer, Heidelberg (2016). \doi{10.10007/1234567890}

% \bibitem{ref_book1}
% Author, F., Author, S., Author, T.: Book title. 2nd edn. Publisher,
% Location (1999)

% \bibitem{ref_proc1}
% Author, A.-B.: Contribution title. In: 9th International Proceedings
% on Proceedings, pp. 1--2. Publisher, Location (2010)

% \bibitem{ref_url1}
% LNCS Homepage, \url{http://www.springer.com/lncs}. Last accessed 4
% Oct 2017
% \end{thebibliography}
\end{document}